\documentclass[twocolumn, switch]{article} 

\usepackage{preprint}

\usepackage{amsmath, amsthm, amssymb, amsfonts}

\usepackage[numbers,square]{natbib}
\usepackage[utf8]{inputenc}	
\usepackage[T1]{fontenc}	
\usepackage{xcolor}		
\usepackage[colorlinks = true,
            linkcolor = purple,
            urlcolor  = blue,
            citecolor = cyan,
            anchorcolor = black]{hyperref}	
\usepackage{booktabs} 		
\usepackage{nicefrac}		
\usepackage{microtype}		
\usepackage{lineno}		
\usepackage{float}			
\usepackage{graphicx}		

\usepackage{newfloat}
\DeclareFloatingEnvironment[name={Supplementary Figure}]{suppfigure}
\usepackage{sidecap}
\sidecaptionvpos{figure}{c}

\usepackage{titlesec}
\titlespacing\section{0pt}{12pt plus 3pt minus 3pt}{1pt plus 1pt minus 1pt}
\titlespacing\subsection{0pt}{10pt plus 3pt minus 3pt}{1pt plus 1pt minus 1pt}
\titlespacing\subsubsection{0pt}{8pt plus 3pt minus 3pt}{1pt plus 1pt minus 1pt}

\usepackage{tikz,xcolor,hyperref}

\definecolor{lime}{HTML}{A6CE39}
\DeclareRobustCommand{\orcidicon}{
	\begin{tikzpicture}
	\draw[lime, fill=lime] (0,0)
	circle [radius=0.16]
	node[white] {{\fontfamily{qag}\selectfont \tiny ID}};
	\draw[white, fill=white] (-0.0625,0.095)
	circle [radius=0.007];
	\end{tikzpicture}
	\hspace{-2mm}
}
\foreach \x in {A, ..., Z}{\expandafter\xdef\csname orcid\x\endcsname{\noexpand\href{https://orcid.org/\csname orcidauthor\x\endcsname}
			{\noexpand\orcidicon}}
}

\title{CAD-feature enhanced machine learning for manufacturing effort estimation on sheet metal bending parts}

\usepackage{authblk}

\author[1,2]{Matteo Ballegeer$^{\dagger}$\orcidA{}}
\author[1,2]{Toon Van Camp$^{\dagger}$\orcidB{}}
\author[3]{Willem Jaspers$^{*}$\orcidC{}}
\author[4]{Alp Bayar\orcidD{}}
\author[4]{Aung Nyein Soe\orcidE{}}
\author[3]{Martin Roelfs\orcidF{}}
\author[1,2]{Dries F. Benoit\orcidG{}}
\author[3]{Bieke Decraemer\orcidH{}}
\author[4]{Joost R. Duflou\orcidI{}}

\affil[1]{Data Analytics Research Group, Ghent University\\9000 Ghent, Belgium}
\affil[2]{Corelab CVAMO, FlandersMake@UGent, Ghent University\\9000 Ghent, Belgium}
\affil[3]{Corelab CodesignS, Flanders Make\\Oude Diestersebaan 133, 3920 Lommel, Belgium}
\affil[4]{Department of Mechanical Engineering, KU Leuven/Flanders Make\\Celestijnenlaan 300, 3001 Leuven, Belgium}

\begin{document}

\twocolumn[ 
  \begin{@twocolumnfalse} 

\maketitle
\vspace{-0.2cm}
\noindent{\footnotesize $^{\dagger}$These authors contributed equally to this work.}
\\
\noindent{\footnotesize $^{*}$Corresponding author: willem.jaspers@flandersmake.be}

\begin{abstract}
Graph-based machine learning has emerged as a promising approach for manufacturability analysis by learning directly from CAD models represented as Boundary Representations (B-reps), exploiting both surface geometry and topological connectivity. However, purely geometric representations often lack the process-specific semantics required for accurate manufacturability prediction: many manufacturing factors, such as surface roles or bend intent, are not explicitly encoded in shape alone and are difficult for data-driven models to infer reliably. We propose a hybrid approach that addresses this challenge by enriching B-rep attributed adjacency graphs with manufacturing features recognized through a rule-based module. Applied to sheet metal bending, recognized features, such as bend characteristics, flange lengths, and surface roles are integrated as node attributes, concentrating the learning signal on process-relevant geometric patterns. Experiments on both a large-scale synthetic manufacturability benchmark and a real-world industrial dataset with measured bending times, one of the first such validations on genuine production data, demonstrate that combining domain knowledge with graph-based learning improves prediction accuracy across both tasks. The results demonstrate that hybrid modeling offers a feasible and effective path toward deployable tools for manufacturability assessment and effort estimation in industrial CAD environments.
\end{abstract}

\noindent\textbf{Keywords:} Manufacturing effort estimation; CAD learning; feature recognition; sheet metal design
\vspace{0.35cm}

  \end{@twocolumnfalse} 
] 


\section{Introduction}

Design for Manufacturing (DFM) aims to identify production constraints early in the design phase to ensure manufacturability, reduce complexity and cost, and preserve functional intent~\cite{anderson2020design}. When manufacturability considerations are neglected, time and resources are often wasted, leading to suboptimal designs or the need to substantially revise or restart the design process~\cite{patterson2021generation}. Automated decision-support tools that integrate DFM guidelines into CAD environments can mitigate these challenges by guiding designers toward manufacturable solutions from the outset~\cite{ferrer2010methodology}.

Recent advances in deep learning have shown that implicit manufacturing knowledge can be learned directly from historical design data~\cite{wang2022machine}. Boundary representations (B-reps), the native format of CAD models, are particularly valuable in this context. A B-rep encodes a solid object as a collection of parametric surfaces connected through shared edges and vertices, naturally inducing a graph in which faces correspond to nodes and shared boundaries to edges. The resulting face adjacency graph~\cite{joshi1988graph} has become the dominant representation for learning from CAD models, as it preserves topological structure without conversion to lossy formats such as meshes or voxels.

Graph neural networks (GNNs) operating on these graphs can exploit both local geometry and global part structure, making them well-suited for two complementary DFM tasks: manufacturability assessment (determining whether a part can be produced within process constraints) and effort estimation (quantifying the time or cost required to do so). Despite strong results on feature recognition benchmarks, their application to part-level characteristics such as manufacturability, effort or cost estimation remains underexplored~\cite{zhang2022novel}. A central challenge is data scarcity: large, high-quality industrial CAD datasets are rare due to intellectual property constraints~\cite{stjepandic2015concurrent,wang2018deep}. Purely data-driven B-rep methods are therefore often difficult to deploy, while rule-based DFM tools, though more data-efficient, are rigid and labour-intensive to maintain~\cite{favi2021cad}.

This work addresses this gap through a hybrid approach that integrates rule-based sheet metal feature recognition with graph-based deep learning. Recognized manufacturing features are injected as node attributes into the B-rep face adjacency graph, allowing the model to jointly reason over geometric structure and manufacturing semantics. This approach concentrates the learning signal on manufacturability-relevant patterns. The approach is evaluated on both a large-scale synthetic dataset and a real-world industrial dataset of measured bending times---one of the first validations of B-rep learning for manufacturing effort on genuine industrial production data.

The remainder of this paper is structured as follows. Section~\ref{sec:related} reviews related work. Section~\ref{sec:method} presents the methodology. Section~\ref{sec:experiments} reports experimental results, and Section~\ref{sec:conclusion} provides the conclusions, limitations, and directions for future work.

\section{Related work}
\label{sec:related}

\subsection{Sheet metal feature recognition}

In the context of manufacturing, feature recognition denotes the extraction of manufacturing-relevant information from CAD models, typically by reasoning over relationships between multiple surfaces and edges. Rule-based approaches have a long history in sheet metal. Early work by Gupta and Gurumoorthy~\cite{gupta2013classification} defined basic deformation features such as bends and flanges, representing compound features through a graph-based formalism operating directly on B-rep models.

Subsequent systems extended this toward richer feature vocabularies and process-specific constraints: Kannan and Shunmugam~\cite{kannan2009processing} developed a STEP-based system using center-plane models with topological rules to identify forming features, while Liu et al.~\cite{liu2025manufacturability} classified faces into web, thickness, bending, and flange categories to recognize bending and concave features in aircraft sheet metal. Ghaffarishahri and Rivest~\cite{ghaffarishahri2020feature} introduced a comprehensive method for aerospace sheet metal that classifies B-rep elements into novel subtypes to recognize flanges, joggles, beads, and their hierarchical relationships, and Slyadnev~\cite{slyadnev2026bending} extended recognition toward process planning through a simulation framework that extracts bend properties from attributed adjacency graphs for feasibility checking.

Deep learning has more recently been applied to sheet metal feature recognition. Ma and Yang~\cite{ma2024adaptive} developed a synthetic dataset with labeled sheet metal bending features and demonstrated learning-based recognition, establishing a benchmark for data-driven approaches. However, deep learning for sheet metal feature recognition remains relatively nascent compared to domains such as machining, and its integration with downstream manufacturability or effort estimation tasks has received little attention.

\subsection{Manufacturability and cost/effort estimation}

Traditional manufacturability estimation compares extracted feature parameters against process-specific rules. Liu et al.~\cite{liu2025manufacturability} evaluate bending radius compliance and concave dimension ratios for sheet metal; Kumar et al.~\cite{kumar2016feature} similarly assess sheet metal features through rule-based methods. These approaches are effective in well-defined contexts but are rigid, labour-intensive to maintain, and provide little guidance on production effort beyond binary feasibility judgements~\cite{favi2021cad}.

Data-driven methods for sheet metal manufacturability and effort estimation remain scarce. Seibold et al.~\cite{seibold2022process} assessed coatability for powder-coated sheet metal by transforming CAD representations into 2D raster formats and processing with Convolutional Neural Networks (CNNs). Beyond sheet metal, Yoo and Kang~\cite{yoo2021explainable} predict manufacturing cost from voxelized CAD models using 3D CNNs, while Zhang et al.~\cite{zhang2022novel} construct machining feature graphs enriched with geometric and precision information to estimate CNC manufacturing cost using GNNs. These studies collectively show that cost and effort prediction from CAD is a tractable learning problem across a range of representations and processes. Our prior work~\cite{ballegeer2026bendfm} introduced a large synthetic sheet metal bending dataset and showed that GNNs on B-rep graphs can predict tooling collisions and several notions of bending effort, while also identifying that the current state-of-the-art leaves meaningful room for improvement. No prior work has used B-rep learning to predict sheet metal bending time derived from real production data.

\subsection{Hybrid expert-knowledge machine learning}

Hybrid approaches that combine rule-based domain knowledge with data-driven learning have been explored across several manufacturing domains. In additive manufacturing, Ko et al.~\cite{ko2021machine} combined decision trees with knowledge graphs for design rule construction, and Aljabali et al.~\cite{aljabali2025generalized} integrated process thresholds with a decision tree classifier for Design for Additive Manufacturing (DfAM) compliance. These works illustrate a common pattern: rule-based knowledge is used to structure or constrain the learning problem, improving data efficiency and making model behaviour more interpretable. The closest approach to ours is that of Zhang et al.~\cite{zhang2022novel}, who construct attribute graphs from rule-recognized machining features enriched with geometric and precision information for GNN-based cost estimation of CNC parts. Their work demonstrates the value of embedding domain knowledge directly into the graph representation. However, their approach operates on a specialized graph consisting only of the recognized features, discarding the raw B-rep geometry.

In contrast, the present work retains the full B-rep graph and enriches it with recognized features as additional node attributes, allowing the model to jointly reason over pure geometric variation and discrete manufacturing semantics. To our knowledge, no prior work has integrated recognized manufacturing features directly into B-rep graph representations in this way.

\subsection{Real-world industrial data validation}

A recurring limitation of the existing literature on CAD-based manufacturability and effort prediction is the near-exclusive reliance on synthetic datasets. This is understandable given the difficulty of obtaining industrial CAD data under intellectual property and confidentiality constraints~\cite{stjepandic2015concurrent} but it raises an important question of generalization: models trained and evaluated on synthetic geometries may not transfer reliably to the complexity and variability of real production environments.

Among the works discussed above, none validates effort or cost prediction on a real-world industrial dataset with measured production times in the sheet metal bending domain. Seibold et al.~\cite{seibold2022process} use industrial data for coatability assessment, but their target is binary rather than a continuous effort metric, and the process differs substantially from bending. Zhang et al.~\cite{zhang2022novel} work with real CNC parts but focus on machining rather than sheet metal. The present study addresses this gap directly by validating on both a large-scale synthetic dataset and a proprietary industrial dataset of sheet metal bending designs annotated with bending times measured on the production floor, providing evidence of practical applicability that is largely absent from the existing literature.

\section{Methodology}
\label{sec:method}

We adopt a hybrid methodology that combines a rule-based feature recognition module with a graph-based deep learning model. The rule-based component extracts manufacturing-relevant information directly from the native STEP representation, after which the graph representation incorporates both geometric and manufacturing descriptors. This integration enhances the downstream learning model with explicit domain knowledge, addressing the limitations of purely data-driven geometry encoding.

\subsection{Feature recognition module}

The feature recognition module operates directly on the STEP file. This choice is motivated by sheet metal bending's surface-level geometric characteristics, where features and relationships of interest are primarily on surfaces rather than volumetric details. Consequently, native STEP format provides greater efficiency and industrial relevance compared to mesh-based approaches used in other processes such as injection moulding~\cite{salaets2022flexible}.

The module recognizes various features and related properties. It first searches for geometric features such as cylindrical surfaces (including radius, axial length, arc angle and convexity as properties) that make abstraction of the actual surface definition (e.g., cylinder, B-spline, NURBS) in the STEP file. This first stage also includes a labelling of all faces as either top, bottom or side, to contextualize which ``side'' of the sheet it occupies in the folded geometry.

The second recognition stage augments this information to find more classical manufacturing features, like bends, holes, side holes (usually unwanted in sheet metal) and the outer contour of the sheet. The bend feature includes common properties: inner and outer radius, length, bend angle, global orientation (+ or $-$, indicating same or opposite fold direction) and several aggregates (min, median, max, mean and std) of the flange length on either side of the bend.

\begin{figure}[t]
  \centering
  \includegraphics[width=\linewidth]{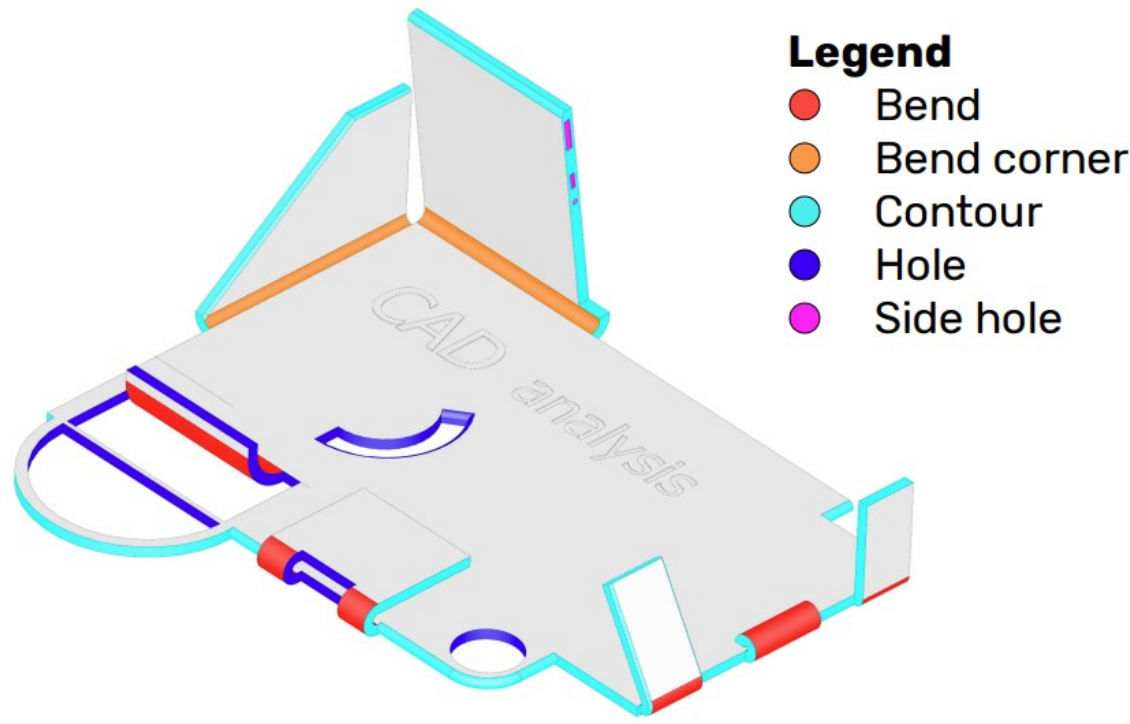}
  \caption{Example feature recognition result.}
  \label{fig:fig1}
\end{figure}

The third recognition stage searches for relational features. In the current implementation, this stage focuses on bend corners, which flags bends that meet or intersect with one another. Figure~\ref{fig:fig1} shows the results of the feature recognition on an example design.

The feature recognition module was designed and developed to be robust. An automatic validation on the BenDFM dataset~\cite{ballegeer2026bendfm} shows a 100\% accuracy for recognition of bend surfaces and their bend properties (angle, radius, length). This demonstrates reliable feature extraction that can be confidently integrated into the downstream graph representation and learning pipeline.

\subsection{Manufacturing feature enhanced graph}

Topology alone is insufficient for capturing geometric relationships in 3D space. The geometry of each surface and its relationship to neighbouring faces must also be encoded. This is achieved by augmenting the graph with node and edge feature vectors, yielding an attributed adjacency graph (AAG)~\cite{cao2020graph}. UV-Net~\cite{jayaraman2021uv} established the dominant paradigm by sampling point coordinates and surface normals in the parametric UV domain of each face and encoding them with a CNN, providing rich descriptors of local surface shape, position, and orientation. FoV-Net~\cite{ballegeer2026fovnet} replaces UV sampling with ray casting and local reference frames, yielding rotation-invariance that has shown improved data efficiency when CAD designs are in varying orientations. Both approaches transform low-level geometric features into 64-dimensional embeddings at the face level.

In addition to these geometric descriptors, our proposed architecture enhances the AAG with manufacturing features recognized by the rule-based module described in Section~\ref{sec:method}. Each face is annotated with boolean indicators for bends, sheet side, holes, and cylindrical surfaces, as well as continuous parameters describing cylindrical geometry (radius, length, angle). Furthermore, detailed bend statistics, including flange orientations and flange length distributions, are incorporated. Together, these manufacturing features yield a 28-dimensional vector per face, which is concatenated with the geometry embedding. Consequently, every node in the graph is represented by a 92-dimensional feature vector that serves as input to the GNN layers.

\begin{figure*}[t]
  \centering
  \includegraphics[width=\linewidth]{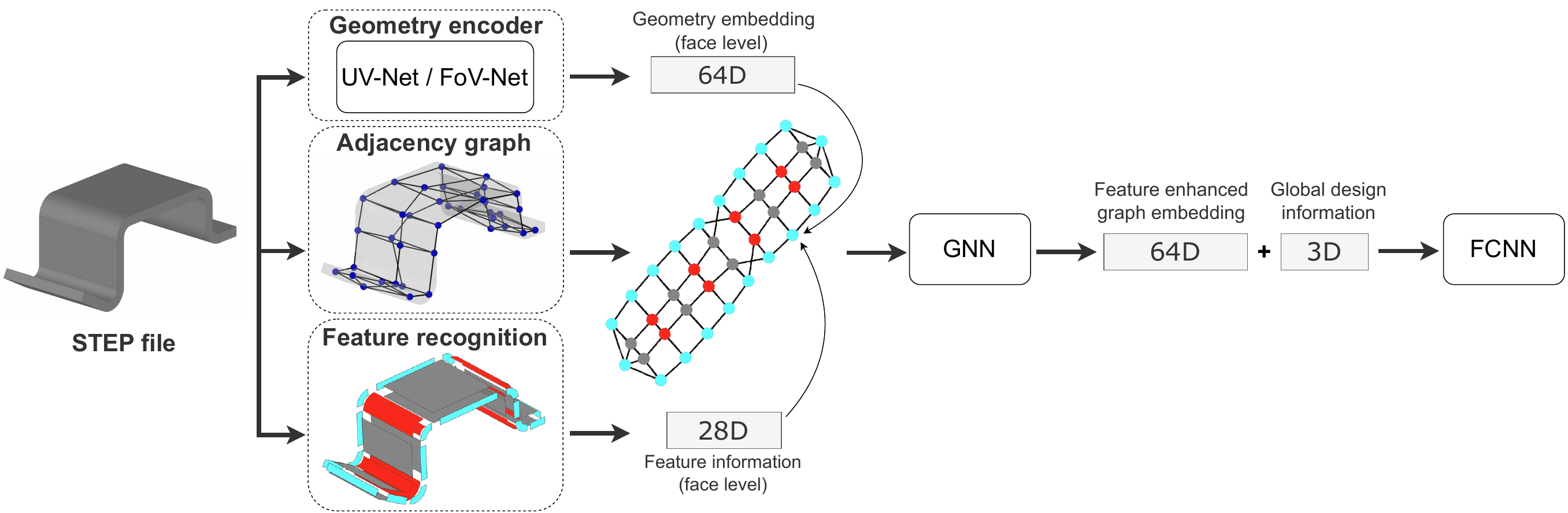}
  \caption{Overview of the proposed architecture.}
  \label{fig:fig2}
\end{figure*}

The GNN leverages the message passing and pooling operations following the architecture of UV-Net and FoV-Net. Message passing allows each face to exchange information with neighboring faces, while pooling aggregates all face embeddings into a single graph-level representation. This results in a 64-dimensional graph-level embedding that captures combined geometric and manufacturing characteristics of the part. Finally, global design attributes (sheet thickness, total face area, and bounding box volume) are appended as a 3-dimensional vector before feeding the representation into a fully connected neural network (FCNN). The FCNN head consists of three layers and outputs either a regression value or class probabilities, depending on the downstream task. Figure~\ref{fig:fig2} illustrates an overview of the full pipeline, from STEP input and feature recognition to the enriched AAG and graph-level prediction.

\section{Experiments}
\label{sec:experiments}

\subsection{Datasets}

The proposed methodology is evaluated on two complementary datasets that cover both controlled benchmarking and real-world applicability.

\textbf{KUL-bend} is a proprietary industrial dataset containing 503 sheet metal bending designs, each associated with a target bending time derived from a timing study conducted on the production floor. Target values correspond to the median observed bending time per part across multiple production instances, reducing the influence of operator variability and measurement noise. The dataset exhibits variability in both design and production characteristics. Parts contain between 1 and 30 bending operations with bend lengths ranging from 12 to 2,680~mm. Sheet thickness varies between 1 and 5~mm (median 2.0~mm), while part areas span six orders of magnitude (1,418 to 8.6M~mm$^2$). Observed bending times range from 3 to 430~seconds (median 49~s, mean 69.6~s, standard deviation 59.7~s). KUL-bend reflects authentic manufacturing environments with realistic heterogeneity in both design geometries and production conditions.

\textbf{BenDFM}~\cite{ballegeer2026bendfm} is a large-scale synthetic dataset consisting of 14,000 STEP files of sheet metal bending designs annotated with manufacturability labels for tooling collisions and handling effort. The scale and label diversity of this dataset make it well suited for evaluating model performance under sufficient data conditions. In this study, we predict the presence of a collision between the part and the tooling (bend or die) during the bending process, formulated as a binary classification task. The dataset is perfectly balanced between the two classes.

\subsection{Models}

To evaluate the contribution of manufacturing feature enrichment, we benchmark two B-rep learning models and their manufacturing-feature-enhanced counterparts.

UV-Net~\cite{jayaraman2021uv} and FoV-Net~\cite{ballegeer2026fovnet} serve as pure B-rep learning baselines, taking only geometric descriptors derived from the STEP file as input. FoV-Net has demonstrated improved performance over UV-Net owing to its rotation invariance, making it the stronger of the two baselines. UV-Net~(MF) and FoV-Net~(MF) extend these baselines by incorporating manufacturing features from the feature recognition module, as described in Section~\ref{sec:method}. We additionally include a Baseline that predicts the training set mean for all test instances (a median bending time of 69.53~s), serving as an important lower bound performance. For the classification task, the baseline is the class distribution (50/50).

\subsection{Experimental setup}

Both datasets are partitioned into fixed training, validation, and test splits. For BenDFM, the official splits are used, while KUL-bend is divided into 80/10/10 proportions. All models are trained using early stopping, terminating training when the validation loss shows no improvement for 50 consecutive epochs. Optimization is performed using the Adam optimizer with a learning rate of 0.0001 and a batch size of 32. UV-Net and FoV-Net follow the implementations described in their respective original publications.

For BenDFM, performance is evaluated using binary classification accuracy. For KUL-bend, we report Root Mean Squared Error (RMSE), Mean Absolute Error (MAE), and Mean Absolute Percentage Error (MAPE). RMSE and MAE measure prediction error in the same unit as the target variable, with RMSE placing greater emphasis on large deviations. MAPE provides a scale-independent metric that facilitates comparison across parts with varying bending times.

\subsection{Results}

Table~\ref{tab:kul} reports the bending time regression results on the KUL-bend dataset. Using the training set mean as a baseline yields an RMSE of 49.8~s, an MAE of 36.8~s, and a MAPE of 106.65\%, highlighting the data's variability and indicating that simple mean-based predictions are insufficient.

Among the pure B-rep learning approaches, FoV-Net consistently outperforms UV-Net across all metrics. This observation reinforces the findings reported in the FoV-Net study and further highlights the importance of rotation invariance in industrial CAD applications. In real-world datasets such as KUL-bend, models cannot rely on consistent canonical orientations of parts, making rotation-robust representations particularly important.

\begin{table}[H]
  \caption{Bending time regression results on KUL-bend. Best results are highlighted in bold. Reported values are mean $\pm$ std.\ over five runs with different random seeds.}
  \label{tab:kul}
  \centering
  \small
  \begin{tabular*}{\linewidth}{@{\hspace{0.05\linewidth}}@{\extracolsep{\fill}}lccc@{\hspace{0.05\linewidth}}}
    \toprule
    Architecture & RMSE (s) & MAE (s) & MAPE (\%) \\
    \midrule
    Baseline              & 49.8             & 36.8             & 106.65 \\
    UV-Net                & $44.50\pm 1.79$  & $28.62\pm 1.67$  & $56.33\pm 10.6$ \\
    UV-Net (MF)           & $42.92\pm 1.27$  & $24.98\pm 0.83$  & $\mathbf{43.86\pm 3.04}$ \\
    FoV-Net               & $35.10\pm 2.16$  & $23.42\pm 0.83$  & $51.14\pm 5.12$ \\
    \textbf{FoV-Net (MF)} & $\mathbf{31.16\pm 1.68}$ & $\mathbf{20.38\pm 1.14}$ & $44.53\pm 5.40$ \\
    \bottomrule
  \end{tabular*}
\end{table}

Augmenting the models with manufacturing feature (MF) information consistently improves performance for both architectures. For UV-Net, the MAE decreases by 3.64~s and the MAPE improves by 12.47 percentage points. FoV-Net shows similar improvements, with a reduction of 3.04~s in MAE and 6.61 percentage points in MAPE. These results indicate that incorporating explicit manufacturing knowledge provides complementary information that benefits learning-based CAD analysis. Overall, FoV-Net~(MF) achieves the best predictive accuracy, obtaining the lowest RMSE and MAE. UV-Net~(MF), however, yields the lowest MAPE.

Table~\ref{tab:bendfm} reports the classification results on the BenDFM dataset. Both B-rep learning architectures substantially outperform the baseline, demonstrating that geometric representations extracted from CAD models provide strong signals for detecting bending collisions. Consistent with the regression results, FoV-Net achieves higher accuracy than UV-Net, highlighting the advantages of rotation-invariant representations when learning directly from CAD geometry. The inclusion of manufacturing feature information yields consistent improvements for both architectures, although the gains are smaller than those observed on the KUL-bend regression task. UV-Net improves from 75.21\% to 76.54\%, while FoV-Net increases from 80.60\% to 81.26\%, with FoV-Net~(MF) achieving the best performance. The smaller improvement may be explained by the substantially larger dataset size (11,200 training samples), which enables the models to learn many manufacturability-relevant patterns directly from geometric information. It may also indicate that the manufacturing descriptors are less relevant for this collision detection task, or that much of their information is already captured by the geometric features.

\begin{table}[t]
  \caption{Collision detection classification results on BenDFM. Best results are highlighted in bold. Reported values are mean $\pm$ std.\ over five runs with different random seeds.}
  \label{tab:bendfm}
  \centering
  \begin{tabular}{lc}
    \toprule
    Architecture & Accuracy (\%) \\
    \midrule
    Baseline              & 50 \\
    UV-Net                & $75.21\pm 0.49$ \\
    UV-Net (MF)           & $76.54\pm 0.49$ \\
    FoV-Net               & $80.60\pm 0.53$ \\
    \textbf{FoV-Net (MF)} & $\mathbf{81.26\pm 0.42}$ \\
    \bottomrule
  \end{tabular}
\end{table}

\section{Conclusion}
\label{sec:conclusion}

This paper presented a hybrid approach for manufacturing effort estimation of sheet metal designs that combines graph-based machine learning with rule-based feature recognition. By integrating recognized manufacturing features into a B-rep graph representation, the proposed method augments geometric information with process semantics that are not directly encoded in geometry alone. Experimental results show that combining geometric and feature-level information improves prediction accuracy compared to approaches relying solely on geometric descriptors. These results highlight the potential of hybrid geometric--semantic representations for enhancing manufacturability prediction in industrial CAD environments.

Several limitations of the current study warrant discussion. First, the KUL-bend dataset, while industrially authentic, is limited to a single production facility and bending process, and generalization to other facilities or process variants remains to be validated. Second, bending time as measured in a timing study captures operator and setup variability that is not encoded in the CAD model itself, introducing irreducible noise in the regression target. Future work should investigate evaluating larger and more diverse industrial datasets and exploring whether the hybrid enrichment strategy transfers to other manufacturing processes such as milling or injection molding.

\section*{Acknowledgements}

This research was supported by Flanders Make, the strategic research center for the Flemish manufacturing industry.

\normalsize
\bibliography{references}

\end{document}